\documentclass[11pt,letterpaper]{article}
\usepackage{emnlp2017}
\usepackage{times}
\usepackage{textcomp}
\usepackage{latexsym}
\usepackage{amssymb}
\usepackage{color}
\usepackage{graphicx}
\usepackage{balance}
\usepackage[font=small]{caption}
\usepackage[shortlabels]{enumitem}

\emnlpfinalcopy

\def\emnlppaperid{1281}

\newcommand{\todo}[1]{}
\renewcommand{\todo}[1]{{\color{red} TODO: {#1}}}
\newcommand{\kgram}[1]{{\it #1}-gram}
\newcommand{\kgrams}[1]{{\it #1}-grams}
\newcommand{\ngram}{\kgram{n}}
\newcommand{\ngrams}{\kgrams{n}}

\newcommand{\denselist}{\setlength{\itemsep}{1pt}
  \setlength{\parskip}{0pt} \setlength{\parsep}{0pt}}
\newcommand{\bitem}{\begin{itemize}[noitemsep,topsep=2pt]\denselist}
\newcommand{\eitem}{\end{itemize}}

\newcommand{\maincaptionref}{\ref{tab:hyperSegmentation}}

\title{Natural Language Processing with Small Feed-Forward Networks}

\author{%
  Jan A. Botha \quad Emily Pitler \quad Ji Ma \quad Anton Bakalov \\[0.8ex]
  {\bf Alex Salcianu} \quad {\bf David Weiss} \quad {\bf Ryan McDonald} \quad {\bf Slav Petrov} \\[0.5ex]
  Google Inc. \\
  Mountain View, CA \\
  {\small \tt \{jabot,epitler,maji,abakalov,salcianu,djweiss,ryanmcd,slav\}@google.com}}

\date{}

\begin{document}

\maketitle

\begin{abstract}
We show that small and shallow feed-forward neural networks can achieve near state-of-the-art results on a range of unstructured and structured language processing tasks while being considerably cheaper in memory and computational requirements than deep recurrent models.
  Motivated by resource-constrained environments like mobile phones, we showcase simple techniques for obtaining such small neural network models, and investigate different tradeoffs when deciding how to allocate a small memory budget.
  \end{abstract}

\section{Introduction}
Deep and recurrent neural networks with large network capacity
have become increasingly accurate for challenging language processing tasks.
For example, machine translation models have been able to attain impressive
accuracies, with models that use hundreds of millions \cite{bahdanau2014neural,googlenmt} or billions \cite{shazeer2017outrageously} of parameters.
These models, however, may not be feasible in all computational settings.
In particular, models running on mobile devices are often constrained
in terms of memory and computation.

Long Short-Term Memory (LSTM) models \cite{hochreiter1997} 
have achieved good results with small memory footprints 
by using character-based input representations:
e.g., the part-of-speech tagging models of \newcite{gillick2016brnn}
have only roughly 900,000 parameters.
Latency, however, can still be an issue with LSTMs, due to the large number of 
matrix multiplications they require (eight per LSTM cell):
\newcite{kimrush2016} report speeds of only 8.8 words/second when running a two-layer LSTM translation system on an Android phone.

Feed-forward neural networks have the potential to be much faster.
In this paper, we show that small feed-forward networks 
can achieve results at or near the state-of-the-art
on a variety of natural language processing tasks,
with an order of magnitude speedup over an LSTM-based approach.

We begin by introducing the network model structure and the character-based
representations we use throughout all tasks (\S\ref{sec:network}).
The four tasks that we address are: language identification (Lang-ID), part-of-speech (POS) tagging, word segmentation, and preordering for translation.
In order to use feed-forward networks for structured prediction tasks, 
we use transition systems \cite{titov2007fast,titov2010latent} with feature embeddings as proposed by \newcite{chen-manning:2014:EMNLP},
and introduce two novel transition systems for the last two tasks.
We focus on \emph{budgeted} models
and ablate four techniques (one on each task)
for improving accuracy for a given memory budget:
\begin{enumerate}[topsep=3pt,itemsep=0pt]
\item Quantization: Using more dimensions and less precision (Lang-ID: \S\ref{sec:langid}).
\item Word clusters: Reducing the network size to allow for word clusters and derived features (POS tagging:  \S\ref{sec:approxmap}).
\item Selected features: Adding explicit feature conjunctions (segmentation:  \S\ref{sec:bigrams}).
\item Pipelines: Introducing another task in a pipeline and allocating parameters to the auxiliary task instead (preordering:  \S\ref{lbl:reordering-task}). 
\end{enumerate}
We achieve results at or near state-of-the-art with small ($< 3$ MB) models on all four tasks.

\section{Small Feed-Forward Network Models}
\label{sec:network}
\begin{figure}[t]
  \centering
  \includegraphics[width=0.9\columnwidth]{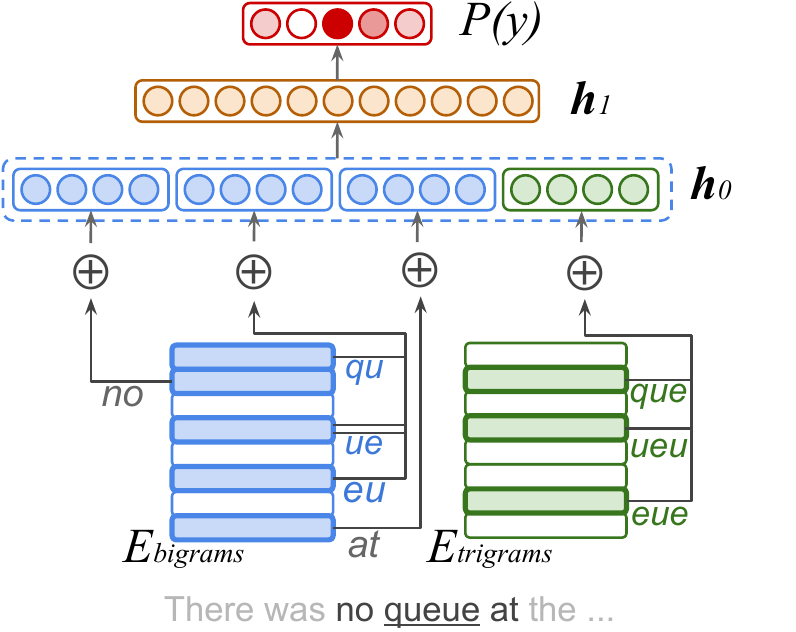}
 \caption{An example network structure for a model using bigrams of the previous, current and next word, and trigrams of the current word. Does not illustrate hashing.
  }
  \label{fig:model}
\end{figure}

The network architectures are designed to
limit the memory and runtime of the model.
Figure \ref{fig:model} illustrates the model architecture:
\begin{enumerate}[topsep=3pt,itemsep=0pt]  
\item Discrete features are organized into groups (e.g., $\mathbf{E}_{\textit{bigrams}}$), with one embedding matrix $\mathbf{E}_g \in \mathbb{R}^{V_g \times D_g}$ per group.
\item Embeddings of features extracted for each group are reshaped into a single vector and concatenated to define the output of the embedding layer as
$\mathbf{h}_0 = [\mathbf{X}_g \mathbf{E}_g \;|\; \forall g ]$.
\item A single hidden layer, $\mathbf{h}_1$, with $M$ rectified linear units \cite{relu} is fully connected to $\mathbf{h}_0$.
\item A softmax function models the probability of an output class y: $P(y) \propto \exp(\mathbf{\beta}_y^T\mathbf{h}_1 + b_y)$, where $\mathbf{\beta}_y \in \mathbb{R}^{M}$ and $b_y$ are the weight vector and bias, respectively.
\end{enumerate}

Memory needs
are dominated by the embedding matrix sizes ($\sum_g V_g D_g$, where $V_g$ and $D_g$ are the vocabulary sizes and dimensions respectively for each feature group $g$),
while runtime is strongly influenced by the
hidden layer dimensions.

\paragraph{Hashed Character \ngrams}
\label{sec:characterngrams}
Previous applications of this network structure used (pretrained) word embeddings to represent words \cite{chen-manning:2014:EMNLP,weiss-etAl:2015:ACL}.
However, for word embeddings to be effective, they usually need to cover large vocabularies (100,000+) and dimensions (50+).
Inspired by the success of character-based representations \cite{ling2015:charscomposed}, we use features defined over character \ngrams{} instead of relying on word embeddings, and learn their embeddings from scratch.

We use a distinct feature group $g$ for each \ngram{} length $N$, and control the size $V_g$ directly by applying \emph{random feature mixing} \cite{ganchev2008small}.
That is, we define the feature value $v$ for an \ngram{} string $x$ as 
$ v=\mathcal{H}(x) \; \mathrm{mod} \; V_g $,
where $\mathcal{H}$ is a well-behaved hash function.
Typical values for $V_g$ are in the 100-5000 range, which is far smaller than the exponential number of unique raw \ngrams{}.
A consequence of these small feature vocabularies is that we can also use small feature embeddings, typically $D_g$=16.

\paragraph{Quantization}
\label{sec:quantization}
A commonly used strategy for compressing neural networks is quantization, using
less precision to store parameters \cite{han2015deep}.
We compress the embedding weights (the vast majority of the parameters for these shallow models) by storing scale factors for each embedding (details in the supplementary material).
In~\S\ref{sec:langid}, we contrast devoting model size to higher precision and lower dimensionality
versus lower precision and more network dimensions.

\paragraph{Training}
Our objective function combines the cross-entropy loss for model predictions relative to the ground truth with L2 regularization of the biases and hidden layer weights.
For optimization, we use mini-batched averaged stochastic gradient descent with momentum \cite{bottou2010sgd,momentum} and exponentially decaying learning rates.
The mini-batch size is fixed to 32 and we perform a grid search for the other hyperparameters, tuning against the task-specific evaluation metric on held-out data, with early stopping.
Full feature templates and optimal hyperparameter settings are given in the supplementary material.

\section{Experiments}
We experiment with small feed-forward networks for 
four diverse NLP tasks: language identification, part-of-speech tagging,
word segmentation, and preordering for statistical machine translation.

\paragraph{Evaluation Metrics}\label{sec:metrics}
In addition to standard task-specific quality metrics, our evaluations also consider model size and computational cost.
We skirt implementation details by calculating size as the number of kilobytes (1KB=1024 bytes) needed to represent all model parameters and resources.
We approximate the computational cost
as the number of floating-point operations (FLOPs) performed for one forward pass through the network given an embedding vector $\mathbf{h}_0$.
This cost is dominated by the matrix multiplications to compute (unscaled) activation unit values,
hence our metric excludes the non-linearities and softmax normalization, but still accounts for the final layer logits.
To ground this metric, we also provide indicative absolute speeds for each task, as measured on a modern workstation CPU (3.50GHz Intel Xeon E5-1650~v3).

\subsection{Language Identification}
\label{sec:langid}
Recent shared tasks on code-switching \cite{molina-codeswitched-task} and dialects \cite{malmasi-EtAl:2016:VarDial3} have generated renewed interest in language identification.
We restrict our focus to single language identification across diverse languages, and
compare to the work of \newcite{baldwin2010language}
on predicting the language of Wikipedia text in 66~languages.
For this task, we obtain the input $\mathbf{h}_0$ by separately \emph{averaging} the embeddings for each \ngram{} length ($N=[1,4]$), as summation did not produce good results.

Table~\ref{tab:quant} shows that
we outperform the low-memory nearest-prototype model of \newcite{baldwin2010language}.
Their nearest neighbor model is the most accurate but its memory scales linearly with the size of the training data.

Moreover, we can apply quantization to the embedding matrix without hurting prediction accuracy:
it is better to use less precision for each dimension, but to use more dimensions.
Our subsequent models all use quantization.
There is no noticeable variation in processing speed when performing dequantization on-the-fly at inference time.
Our 16-dim Lang-ID model runs at 4450~documents/second (5.6~MB of text per second) on the preprocessed Wikipedia dataset.

\paragraph{Relationship to Compact Language Detector}
  These techniques back the open-source Compact Language Detector v3 (CLD3)\footnote{\url{github.com/google/cld3}}
  that runs in Google Chrome browsers.\footnote{As of the date of this writing in 2017.}
Our experimental Lang-ID model uses the same overall architecture as CLD3, but uses a simpler feature set, less involved preprocessing, and covers fewer languages.

\begin{table}[t]
\scalebox{0.9}{
\begin{tabular} {l|cc}
\bf Model & \bf Micro F1 & \bf Size \\
\hline
  \newcite{baldwin2010language}: NN & 90.2 & -\\  
  \newcite{baldwin2010language}: NP & 87.0 & -\\   
  Small FF, 6 dim & 87.3 & 334 KB \\
  Small FF, 16 dim & 88.0 & 800 KB\\
  Small FF, 16 dim, {\it quantized} & 88.0 & 302 KB\\
\end{tabular}}
  \caption{Language Identification.  Quantization allows trading numerical precision for larger embeddings.  The two models from \newcite{baldwin2010language} are the nearest neighbor (NN) and nearest prototype (NP) approaches.}
\label{tab:quant}
\end{table}

\subsection{POS Tagging}
We apply our model as an unstructured classifier to predict a POS tag for each token independently,
and compare its performance to that of the byte-to-span (BTS) model \cite{gillick2016brnn}.
BTS is a 4-layer LSTM network that maps a sequence of bytes to a sequence of labeled spans, such as tokens and their POS tags.
Both approaches limit  model size by using small input vocabularies: byte values in the case of BTS, and 
hashed character \ngrams \ and (optionally) cluster ids in our case.

\paragraph{Bloom Mapped Word Clusters}
\label{sec:approxmap}
It is well known that word clusters can be powerful features in linear models for a variety of tasks \cite{koo2008simple,Turian:2010:wordreps}.
Here, we show that they can also be useful in neural network models.
However, naively introducing word cluster features drastically increases
the amount of memory required,
as a word-to-cluster mapping file with hundreds of thousands of entries
can be several megabytes on its own.\footnote{For example, the commonly
used English clusters from the BLLIP corpus is over 7 MB -- {\footnotesize\url{people.csail.mit.edu/maestro/papers/bllip-clusters.gz}}}
By representing word clusters with a Bloom map \cite{talbot2008bloom},
a key-value based generalization of Bloom filters,
we can reduce the space required by a factor of $\sim$15
and use 300KB to (approximately) represent the clusters for 250,000 word types.

\begin{table}
\scalebox{0.9}{
  \begin{tabular}{l|cccc}
    {\bf Model} & \bf Acc. & \bf Wts. & \bf MB & \bf Ops. \\ \hline
   \newcite{gillick2016brnn} & 95.06 & 900k &  -  & 6.63m \\
    Small FF   & 94.76 & 241k & 0.6 & 0.27m \\
    \,\, +Clusters& 95.56 & 261k & 1.0 & 0.31m \\
    $\qquad \frac{1}{2}$ Dim.
              & 95.39 & 143k & 0.7 & 0.18m \\
\end{tabular}}
  \caption{%
    POS tagging. 
    Embedded word clusters improves accuracy and allows the use of smaller
    embedding dimensions.
    }
\label{tab:brnnpos}
\end{table}

In order to compare against the monolingual setting of \newcite{gillick2016brnn},
we train models for the same set of 13 languages from the Universal Dependency treebanks v1.1 \cite{UD1lrec2016} corpus, using the standard predefined splits.

As shown in Table~\ref{tab:brnnpos},
our best models are 0.3\% more accuate on average across all languages than the BTS monolingual models, while using 6x fewer parameters and 36x fewer FLOPs. 
The cluster features play an important role, providing
a 15\% relative reduction in error over our vanilla model, but also increase the overall size.
Halving all feature embedding dimensions (except for the cluster features)
still gives a 12\% reduction in error
and trims the overall size back to 1.1x the vanilla model, staying well under 1MB in total.
This halved model configuration has a throughput of 46k~tokens/second, on average.

Two potential advantages of BTS are that it does not require tokenized input and has a more accurate multilingual version, achieving 95.85\% accuracy.
From a \emph{memory} perspective, one multilingual BTS model will take less space than
separate FF models.  
However, from a \emph{runtime} perspective, a pipeline of our models doing language identification,
word segmentation, and then POS tagging would still be faster than a single instance of the deep LSTM BTS model, by about 12x in our FLOPs estimate.\footnote{Our calculation of BTS FLOPs is very conservative and favorable to BTS, as detailed in the supplementary material.}

\subsection{Segmentation}
Word segmentation is critical for processing Asian languages where words are not explicitly separated by spaces.
Recently, neural networks have significantly improved segmentation accuracy
\cite{zhang-zhang-fu:2016:P16-1,cai-zhao:2016:P16-1,DBLP:conf/ijcai/LiuCGQL16, DBLP:journals/corr/YangZD17, DBLP:journals/corr/KongDS15}.
  We use a structured model based on the transition system in Table~\ref{tab:segmentationtransitions}, and similar to the one proposed by \newcite{zhang-clark2007}.
We conduct the segmentation experiments on the Chinese Treebank 6.0 with the recommended data splits. No external resources or pretrained embeddings are used.
Hashing was detrimental to quality in our preliminary experiments, hence we do not use it for this task.
To learn an embedding for unknown characters, we cast characters occurring only once in the training set to a special symbol.

\begin{table}[t]
\begin{tabular}{ll}
\bf Transition & \\
\hline
{\sc Split} & $([\sigma], [i|\beta]) \rightarrow ([\sigma | i], [\beta])$  \\
{\sc Merge} & $([\sigma], [i | \beta]) \rightarrow ([\sigma], [\beta])$  \\
\end{tabular}
\caption{Segmentation Transition system.  Initially all characters are on the buffer $\beta$ and the stack $\sigma$ is empty:
$([], [c_1 c_2 ... c_n])$.  In the final state the buffer is empty and the stack contains the first character for each word.}
\label{tab:segmentationtransitions}
\end{table}

\begin{table}[t]
\scalebox{0.9}{
\begin{tabular} {l|cc}
\bf Model & \bf Accuracy & \bf Size \\
\hline
\newcite{zhang-zhang-fu:2016:P16-1} & 95.01  & $-$ \\ 
\newcite{zhang-zhang-fu:2016:P16-1}-combo & 95.95  & $-$ \\
Small FF, 64 dim & 94.24 & 846KB \\
Small FF, 256 dim& 94.16 & 3.2MB \\
Small FF, 64 dim, \it bigrams & 95.18 & 2.0MB \\
\end{tabular}}
\caption{Segmentation results. Explicit bigrams are useful.}
\label{tab:bigrams}
\end{table}

\paragraph{Selected Features}
\label{sec:bigrams}
Because we are not using hashing here, we need to be careful about the size of the input vocabulary.
The neural network with its non-linearity is in theory able to learn bigrams by conjoining unigrams, 
but it has been shown that explicitly using character bigram features leads to better accuracy \cite{zhang-zhang-fu:2016:P16-1,pei-ge-chang:2014:P14-1}.
\newcite{zhang-zhang-fu:2016:P16-1} suggests that embedding manually specified feature conjunctions further improves accuracy (`\newcite{zhang-zhang-fu:2016:P16-1}-combo' in Table \ref{tab:bigrams}).  However, such embeddings could easily lead to a model size explosion and thus are not considered in this work. 

The results in Table~\ref{tab:bigrams} show that spending our memory budget on small bigram embeddings is more effective than on larger character embeddings, in terms of both accuracy and model size.
Our model featuring bigrams runs at 110KB of text per second, or 39k~tokens/second.

\subsection{Preordering}\label{lbl:reordering-task}
Preordering source-side words into the target-side word order is a useful preprocessing task for statistical machine translation \cite{xia2004improving,collinsclause,nakagawa2015efficient,degispert-iglesias-byrne:2015:NAACL-HLT}.
We propose a novel transition system for this task (Table \ref{tab:reorderingtransitions}),
so that we can repeatedly apply a small network to produce these permutations. 
Inspired by a non-projective parsing transition system \cite{nivre:2009:ACL}, the system uses  a {\sc swap} action to permute
spans.
The system is sound for permutations: any derivation will end with all of the input words in a permuted order, and complete: all permutations are reachable (use {\sc shift} and {\sc swap} operations to perform a bubble sort, then {\sc append} $n-1$ times to form a single span).
For training and evaluation, we use the English-Japanese manual word alignments from \newcite{nakagawa2015efficient}.

\begin{table}[t]
\begin{tabular}{ll}
\multicolumn{2}{l}{\bf Transition  \qquad \qquad \qquad \qquad {\bf Precondition}} \\
\hline
{\sc Append} & $([\sigma|i|j], [\beta]) \rightarrow ([\sigma | [i j]], [\beta])$  \\
{\sc Shift} & $([\sigma], [i | \beta]) \rightarrow ([\sigma | i], [\beta])$  \\
{\sc Swap} & $([\sigma|i|j], [\beta]) \rightarrow [\sigma | j], [i | \beta])$; $i < j$\\
\end{tabular}
\caption{Preordering Transition system.  Initially all words are part of singleton spans on the buffer:
$([], [[w_1] [w_2] ... [w_n]])$.  In the final state the buffer is empty and the stack contains a single span.}
\label{tab:reorderingtransitions}
\end{table}

\begin{table}[t]
\scalebox{0.9}{
\begin{tabular} {l|cc}
\bf Model & \bf FRS & \bf Size \\
\hline
\newcite{nakagawa2015efficient} & 81.6 & -\\
Small FF & 75.2 &  0.5MB\\
Small FF + POS tags & 81.3 & 1.3MB \\
Small FF + Tagger input fts. &76.6 & 3.7MB\\
\end{tabular}}
\caption{Preordering results for English $\rightarrow$ Japanese. \emph{FRS} (in $[0, 100]$) is the fuzzy
	reordering score \cite{talbot2011}. }
\label{tab:pipeline}
\end{table}

\paragraph{Pipelines}
\label{sec:pipelines}
For preordering, we experiment with either spending all of our memory budget on reordering, or spending some of the memory budget on features over predicted POS tags, which also requires an additional neural network to predict these tags.
Full feature templates are in the supplementary material.
As the POS tagger network uses features based on a three word window around the token,
another possibility is to add all of the features that would have affected the POS tag of a token to the reorderer directly.

Table \ref{tab:pipeline} shows results with or without using the predicted POS tags in the preorderer,
as well as including the features used by the  tagger in the reorderer directly and only training the downstream task.  
 The preorderer that includes a separate network for POS tagging and then extracts features over the predicted tags is 
 more accurate and smaller than the model that includes all the features that contribute to a POS tag in the reorderer directly.
This pipeline processes 7k~tokens/second when taking pretokenized text as input, with the POS tagger accounting for 23\% of the computation time.

\section{Conclusions}
This paper shows that small feed-forward networks are sufficient to achieve
useful accuracies on a variety of tasks.
In resource-constrained environments, speed and memory are important 
metrics to optimize as well as accuracies.
While large and deep recurrent models are likely to be the most accurate whenever
they can be afforded,
feed-foward networks can provide better value in terms of runtime and memory,
and should be considered a strong baseline.

\section*{Acknowledgments}
We thank Kuzman Ganchev, Fernando Pereira, and the anonymous reviewers for their useful comments.

\balance

\bibliography{emnlp2017}
\bibliographystyle{emnlp_natbib}

\clearpage
\newpage
\appendix
\renewcommand{\thefigure}{\roman{figure}}
\renewcommand{\thetable}{\roman{table}}

\section*{\Large Supplementary Material}
\vspace{1ex}

\section{Quantization Details}
The values comprising a generic embedding matrix \mbox{$\mathbf{E} \in \mathbb{R}^{V \times D}$} are ordinarily stored with \mbox{32-bit} floating-point precision in our implementation.
For quantization, we first calculate a scale factor $s_i$ for each embedding vector $\mathbf{e}_i$ as \[s_i=\frac{1}{b-1}\max_j \left|e_{ij}\right|.\]
Each weight $e_{ij}$ is then quantized into an 8-bit integer as \[q_{ij}=\lfloor\frac{1}{2} + \frac{e_{ij}}{s_i} + b\rfloor,\] where the bias $b=128$.
Hence, the number of bits required to store the embedding matrix is reduced by a factor of 4, in exchange for storing the $V$ additional scale values. 
At inference time, the embeddings are dequantized on-the-fly. 

\section{FLOPs Calculation}\label{sec:flop-detail}
The product of $\mathbf{A} \in \mathbb{R}^{P \times Q}$ and $\mathbf{b} \in \mathbb{R}^Q$ involves \mbox{$P(2Q-1)$} FLOPs, and our single ReLu hidden layer requires performing this operation once per timestep ($P$=$M$, $Q$=$H_0$).
Here, $H_0$ denotes the size of the embedding vector $\mathbf{h}_0$, which equals 408, 464 and 260 for our respective POS models as ordered in Table~\ref{tab:brnnpos}.

In contrast, each LSTM layer requires eight products per timestep, and the BTS model has four layers ($P$=$Q$=320).
The particular sequence-to-sequence representation scheme of \newcite{gillick2016brnn} requires at least four timesteps to produce a meaningful output: the individual input byte(s), and a start, length and label of the predicted span.
A single timestep is therefore a relaxed lower bound on the number of FLOPs needed for BTS inference.

\section{Word Clusters}
The word clusters we use are for the 250k most frequent words from a large unannotated corpus that was clustered into 256 classes using the distributed Exchange algorithm \cite{uszkoreit2008distributed} and the procedure described in Appendix~A of \newcite{tackstrom2012clusters}.

The space required to store them in a Bloom map is calculated using the formula derived by \newcite{talbot2008bloom}: each entry requires
 $1.23*(\log\frac{1}{\epsilon} + H)$ bits, where $H$ is the entropy of the distribution on the set of values, and $\epsilon=2^{-E}$, with $E$ the number of error bits employed.
We use 0~error bits and assume a uniform distribution for the 256 values, i.e.\ $H=8$, hence we need 9.84~bits per entry, or 300KB for the 250k entries.

\section{Lang-ID Details}
In our language identification evaluation, the 1,2,3,4-gram embedding vectors each have 6 or 16 dimensions, depending on the experimental setting.
Their hashed vocabulary sizes ($V_g$) are 100, 1000, 5000, and 5000, respectively. 
The hidden layer size is fixed at $M$=208.

We preprocess data by removing non-alphabetic characters and pieces of markup text (i.e., anything located between $<$ and $>$, including the brackets).
At test time, if this results in an empty string, we skip the markup removal, and if that still results in an empty string, we process the original string.
This procedure is an artefact of the Wikipedia dataset, where some documents contain only punctuation or trivial HTML code, yet we must make predictions for them to render the results directly comparable to the literature.

\section{POS Details}
The Small FF model in the comparison to BTS uses 2,3,4-grams and some byte unigrams (see feature templates in Table~\ref{tab:pos-features}).
The \ngrams{} have embedding sizes of 16 and the byte unigrams get 4 dimensions. In our $\frac{1}{2}$-dimension setting, the aforementioned dimensions are halved to 8 and 2.

Cluster features get embedding vectors of size~8.
The hashed feature vocabularies for \ngrams{} are 500, 200, and 4000, respectively.
The hidden layer size is fixed at $M$=320.

\begin{table}[htbp]
  \begin{tabular}{l|l}
    bytes & $\forall i \in [0,1], \forall j \in [0,3]: \, l_{\pm i}^{\pm j}$ \\ \hline
    char \ngrams & $\forall i \in [0,3], \forall N \in [2,4]: \, \{u_{\pm i}^{(N)}\}$ \\ \hline
    clusters & $\forall i \in [0,3]: \, c_{\pm i}$ \\
  \end{tabular}
  \caption{{\bf POS tagging feature templates.} $i$ is a position relative to the focus token. $l_j$ is the value of the $j$-th UTF8 byte from the start/end of a word. $\{u^{(N)}\}$ designates the set of Unicode character \ngrams \ in a word. $c$ is the cluster id of a word.}
  \label{tab:pos-features}
\end{table}

\begin{table}[htbp]
\begin{tabular}{l|l}
\hline
  char & $\forall i \in [0, 1]:$ $\sigma_{\pm i}.$c; \hspace{0.15cm} $\forall i \in [0, 2]$ $\beta_{\pm i}.$c \\ \hline
  bigram & $\forall i \in [0, 1]:$ $\sigma_{\pm i}.$b; \hspace{0.15cm} $\beta_{\pm i}.$b \\\hline

\end{tabular}
  \caption{{\bf Word segmentation feature templates.} `$\beta_{\pm i}$' denotes starting at the $i$-th character to the left/right of the front of the buffer.  `c' and `b' denote character and character-bigram, respectively.}
\label{tab:seg-feature}
\end{table}

\begin{table}[htbp]
  \begin{tabular}{l|l}
  Features & Positions\\
  \hline
    char bigrams & for $i \in [0,1] \,\sigma(i)_1$  \\ 
                          & for $i \in [0,2] \,\sigma(i)_{l_{\sigma(i)}}$  \\ 
                          & $\beta(0)_1$  \\ 
     bytes & for $i \in [0,1] \,\sigma(i)_1$  \\ 
                          & for $i \in [0,2] \,\sigma(i)_{l_{\sigma(i)}}$  \\ 
                          & $\beta(0)_1$  \\ 
   has-swapped & for $i \in [0,1] \sigma(i)$ \\
    tags-main & for $i \in [0,1] \, \sigma(i)_1$ \\
                     & for $i \in [0,2] \, \sigma(i)_{l_{\sigma(i)}}$ \\
                     & $\beta(0)_1$ \\
   tags-aux & for $i \in [0,1] \, \sigma(i)_2 \, \sigma(i)_{l_{\sigma(i)-1}} $\\
                   & for $i \in [2,3] \, \sigma(i)_1 \; \sigma(3)_{l_{\sigma(3)}} $\\
                   & $\beta(0)_2 \, \beta(0)_{l_\beta(0)-1} \, \beta_{l_{\beta(0)}}$ \\
                   & for $j \in [1,3] \, \beta(j)_1 \, \beta(j)_{l_\beta(j)} $\\               
    
           \hline
  \end{tabular}
  \caption{{\bf Preordering feature templates.} Each feature group applies to the set of positions given. $\sigma(i)$ denotes the $i$-th span from the top of the stack, and $\beta(j)$ the $j$-th span from the front of the buffer.
  Within a span $s$, the $l_s$ tokens are $s_{1} ... s_{l_s}$, so $s_1$ is the leftmost token in $s$ and $s_{l_s}$ the rightmost.}
  \label{tab:preordering-features}
\end{table}

\begin{table}[htbp]
\begin{tabular}{l|rrrrr}
{\it Model.}   & {\it L.R.} & {\it Mom.}  &  \multicolumn{1}{c}{$\gamma$} & {\it Steps} & {\it D.P.} \\ \hline
C-64 & 0.03 & 0.8 & 32K & 3.8M & 0.2 \\
C-256 & 0.03 & 0.8  & 32K & 3.6M &0.4 \\
C-64+B-04 & 0.03 & 0.8 & 64K & 7.6M  &0.3 \\
\end{tabular}
  \caption{{\bf Segmentation:} Optimal hyperparameter settings per model for our segmentation experiments reported in Table~\ref{tab:bigrams}.
  The columns show learning rate (L.R.), momentum factor (Mom.), the step-frequency at which the learning rate is scaled by 0.96 ($\gamma$),
and the number of steps at which training was stopped because accuracy peaked on the held-out tuning data.
  The column {\it D.P.} shows the optimal dropout probability. }
  \label{tab:hyperSegmentation}
\end{table}

\begin{table}[htbp]
\begin{tabular}{l|rrrr}
  & {\it L.R.} & {\it Mom.} & \multicolumn{1}{c}{$\gamma$} & {\it Steps} \\ \hline
  No POS tags & 0.05 & 0.9 & 2k & 38k \\
  w/ POS tags & 0.05 & 0.9 & 8k & 46k \\
  $\qquad$ \emph{POS model} & 0.05 & 0.9 & 8k & 500k \\
  w/ tagger input fts. & 0.1 & 0.8 & 4k & 76k \\
\end{tabular}
  \caption{{\bf Preordering:} Optimal hyperparameter settings obtained for our preordering experiments reported in Table~\ref{tab:pipeline}.
  Columns have the same meanings as in Table~\maincaptionref.
  }
  \label{tab:hyper-reorder}
\end{table}

\section{Segmentation Details}
Feature templates used in segmentation experiments are listed in Table \ref{tab:seg-feature}.
Besides, we define length feature to be the number of characters between top of $\sigma$ and the front of $\beta$, this maximum feature value is clipped to 100.
The length feature is used in all segmentation models, and the embedding dimension is set to 6. 
We set the cutoff for both character and character-bigrams to 2 in order to learn unknown character/bigram embeddings. 
The hidden layer size is fixed at $M$=256.

\section{Preordering Details}
The feature templates for the preorderer look at the top four spans on the stack and the first four spans in the buffer;
for each span, the feature templates look at up to the first two words and last two words  within the span.
The ``vanilla'' variant of the preorderer includes character {\it n}-grams, word bytes, and whether the span has ever participated in a {\sc swap} transition.
The POS features are the predicted tags for the words in these positions.  Table \ref{tab:preordering-features} shows the full feature templates for the preorderer.

\begin{table}[htbp]
\begin{tabular}{r|rrr|rrr}
  & \multicolumn{3}{|l}{\bf Small FF 6 dim} & \multicolumn{3}{|l}{\bf Small FF 16 dim}  \\
  {\it \#} & {\it L.R.} & {\it Mom.} & \multicolumn{1}{c|}{$\gamma$} & 
               {\it L.R.} & {\it Mom.} & \multicolumn{1}{c}{$\gamma$}  \\ \hline
  0 & 0.4 & 0.9 & 8k & 0.4 & 0.9 & 16k \\
  1 & 0.4 & 0.9 & 32k & 0.4 & 0.9 & 32k \\
  2 & 0.4 & 0.9 & 32k & 0.4 & 0.9 & 8k \\
  3 & 0.3 & 0.9 & 64k & 0.5 & 0.9 & 16k \\
  4 & 0.4 & 0.8 & 100k & 0.4 & 0.9 & 32k \\
  5 & 0.5 & 0.8 & 100k & 0.4 & 0.9 & 64k \\
  6 & 0.3 & 0.9 & 32k & 0.3 & 0.9 & 32k \\
  7 & 0.3 & 0.9 & 100k & 0.5 & 0.9 & 16k \\
  8 & 0.4 & 0.9 & 32k & 0.5 & 0.9 & 8k \\
  9 & 0.4 & 0.9 & 32k & 0.3 & 0.9 & 16k \\
\end{tabular}
  \caption{{\bf Lang-ID:} Optimal hyperparameter settings obtained with the results reported in Table~\ref{tab:quant}. The first column is the cross-validation fold, while the other
  columns have the same meanings as in Table~\maincaptionref.
  }
  \label{tab:hyper-langid}
\end{table}

\begin{table}[htbp]
\begin{tabular}{l|rrrrr}
  \multicolumn{1}{l}{\it Lang.}   & {\it L.R.} & {\it Mom.} & \multicolumn{1}{c}{$\gamma$}  & {\it Steps} & {\it Acc.} \\
\multicolumn{4}{l}{\bf Small FF}  &  &  \\ \hline
bg & 0.1 & 0.8 & 32k & 90k & 97.12 \\
cs & 0.05 & 0.9 & 32k & 480k & 97.97 \\
da & 0.05 & 0.9 & 32k & 480k & 94.17 \\
en & 0.01 & 0.9 & 128k & 660k & 92.50 \\
fi & 0.05 & 0.9 & 8k & 210k & 93.84 \\
fr & 0.1 & 0.8 & 64k & 60k & 95.10 \\
de & 0.1 & 0.8 & 8k & 120k & 91.23 \\
el & 0.08 & 0.9 & 64k & 60k & 96.88 \\
id & 0.08 & 0.8 & 32k & 180k & 91.60 \\
it & 0.08 & 0.8 & 128k & 330k & 96.79 \\
fa & 0.08 & 0.9 & 128k & 60k & 95.80 \\
es & 0.1 & 0.8 & 32k & 60k & 94.37 \\
sv & 0.1 & 0.9 & 8k & 210k & 94.54 \\
\multicolumn{4}{l}{\bf Small FF + Clusters }  &  &  \\ \hline
bg & 0.08 & 0.8 & 64k & 120k & 97.72 \\
cs & 0.1 & 0.8 & 16k & 420k & 98.12 \\
da & 0.1 & 0.8 & 32k & 360k & 95.49 \\
en & 0.05 & 0.8 & 8k & 510k & 93.88 \\
fi & 0.1 & 0.8 & 8k & 300k & 94.97 \\
fr & 0.05 & 0.9 & 8k & 630k & 95.65 \\
de & 0.05 & 0.9 & 8k & 480k & 92.40 \\
el & 0.1 & 0.9 & 8k & 60k & 97.60 \\
id & 0.1 & 0.8 & 64k & 150k & 91.94 \\
it & 0.1 & 0.8 & 32k & 270k & 97.36 \\
fa & 0.08 & 0.9 & 64k & 90k & 96.24 \\
es & 0.05 & 0.9 & 128k & 30k & 95.01 \\
sv & 0.08 & 0.9 & 16k & 150k & 95.90 \\
  \multicolumn{4}{l}{\bf Small FF ($\frac{1}{2}$ Dim.) + Clusters}  &  &  \\ \hline
bg & 0.1 & 0.8 & 128k & 210k & 97.76 \\
cs & 0.05 & 0.9 & 32k & 420k & 98.06 \\
da & 0.05 & 0.9 & 16k & 240k & 95.33 \\
en & 0.05 & 0.8 & 8k & 300k & 93.06 \\
fi & 0.05 & 0.9 & 16k & 390k & 94.66 \\
fr & 0.08 & 0.9 & 128k & 120k & 95.28 \\
de & 0.08 & 0.9 & 16k & 90k & 92.13 \\
el & 0.08 & 0.9 & 16k & 60k & 97.42 \\
id & 0.08 & 0.9 & 8k & 690k & 92.15 \\
it & 0.05 & 0.9 & 64k & 210k & 97.42 \\
fa & 0.1 & 0.8 & 8k & 510k & 96.19 \\
es & 0.08 & 0.9 & 8k & 60k & 94.79 \\
sv & 0.1 & 0.8 & 16k & 300k & 95.76 \\
\end{tabular}
  \caption{{\bf POS:} Optimal hyperparameter settings per language obtained for our POS experiments.
  Columns have the same meanings as in Table~\maincaptionref.
  The final column shows the test set accuracies that back the averages shown in Table~\ref{tab:brnnpos}.}
  \label{tab:hyperpos}
\end{table}

\end{document}